\def\year{2020}\relax
\documentclass[letterpaper]{article} 
\usepackage{aaai20}  
\usepackage{times}  
\usepackage{helvet} 
\usepackage{courier}  
\usepackage[hyphens]{url}  
\usepackage{graphicx} 
\usepackage{graphicx}  
\usepackage{algorithm}
\usepackage{algpseudocode}
\usepackage{amsmath}
\usepackage{amsfonts}
\usepackage{footnote}
\usepackage{pifont}
\usepackage{booktabs}
\title{Fast Image Caption Generation with Position Alignment}
\author{Zhengcong Fei\textsuperscript{\rm 1,2}\\ 
\textsuperscript{\rm 1}Key Lab of Intelligent Information Processing of Chinese Academy of Sciences (CAS),\\
Institute of Computing Technology, CAS, Beijing 100190, China\\ 
\textsuperscript{\rm 2}University of Chinese Academy of Sciences, Beijing 100049, China\\
feizhengcong@ict.ac.cn
}
 
\begin{document}

\maketitle

\begin{abstract}
Recent neural network models for image captioning usually employ an encoder-decoder architecture, where the decoder adopts a recursive sequence decoding way.
However, such  autoregressive decoding may result in sequential error accumulation and slow generation which limit the applications in practice. Non-autoregressive (NA) decoding has been proposed to cover these issues but suffers from language quality problem due to the indirect modeling of the target distribution. Towards that end, we propose an improved NA prediction framework to accelerate image captioning. Our decoding part consists of a position alignment to order the words that describe the content detected in the given image, and a fine non-autoregressive decoder to generate elegant descriptions. Furthermore, we introduce an inference strategy that regards position information as a latent variable to guide the further sentence generation. The Experimental results on public datasets show that our proposed model achieves better performance compared to general NA captioning models, while achieves comparable performance as autoregressive image captioning models with a significant speedup.
\end{abstract}

\section{Introduction}

Image captioning aims to automatically describe the content of a given image through natural language \cite{Girish2011Baby,Donahue2015Long}.
Inspired by the development of neural machine translation (NMT), recent approaches to image captioning under a general encoder-decoder framework have achieved great success \cite{Vinyals2015Show,Xu2015Show,Xu2016Guiding,Anderson2017Bottom}. In such a framework, an image encoder which is based on a convolutional neural network (CNN) is first used to extract region-level visual feature vectors for a given image, a caption decoder which is based on a recurrent neural network (RNN) is then adopted to generate caption words recurrently.
Despite their success, the state-of-the-art image caption models usually suffer from the slow inference speed, which has become a bottleneck to apply deep learning-based models in real-time systems \cite{Ranzato2016Sequence}. The slow inference speed of image caption models is due to their autoregressive property, \emph{i.e.}, decoding the target sentence word-by-word according to the given image and generated substance.

\begin{figure}[t]
	\begin{center}
		{\includegraphics[width=0.95\columnwidth]{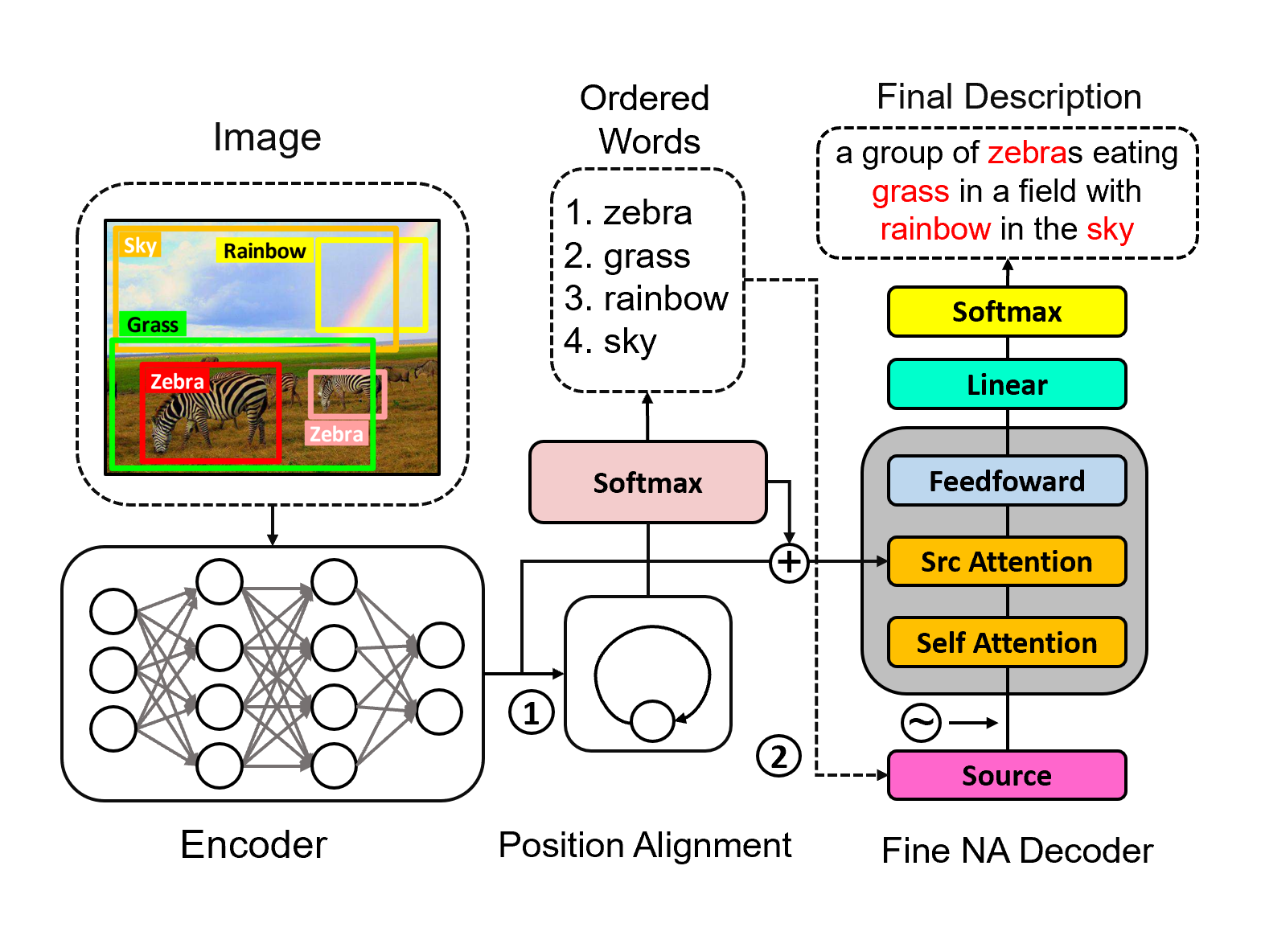}} 
	\end{center}
	{		\caption{
			The framework of our proposed NA captioning model.
			\ding{172} Different from general NA decoding models, our model adds a position alignment module between the image encoder module and sentence decoder module to explicitly construct object and position information. 
			\ding{173} For a general NA model, the decoder inputs are the copied of source information, but for our model, the decoder inputs are ordered words with image information. 
		}
		\label{fig1}}
\end{figure}

Recently, a non-autoregressive (NA) technology \cite{GuNon} has been introduced in NMT which can simultaneously decode all target words to break the bottleneck of the autoregressive generation models. Non-autoregressive sentence generation models usually directly copy the source word representations to the input of the decoder, instead of using previous predicted target word representations \cite{Bingzhen2019Imitation,Guo2018Non}. Hence, the inference of different target words are independent, which enables parallel computation of the sentence decoder.
Non-autoregressive decoding models could achieve 10-15 times speedup compared to autoregressive models while maintaining considerable performance \cite{GuNon}.

However, existing non-autoregressive systems ignore the dependencies among target words and simultaneously generate all target words, which makes the generated sentences have apparent problems in naturalness and fluency. Specifically, when decoding a target word,  the non-autoregressive models will build an indirect distribution of the true target. Consequently, this issue makes non-autoregressive models generate sentences conditioned on less or inaccurate source information, thus leading to missing, repeated and even wrong sentences. 
Moreover, in an image captioning task, the input of an encoder is a given image, which is a low-level semantic data rather than a high-level semantic sequence in NMT.  This brings more difficult for decoder to generation accuracy sentences.

In this paper, to address the above issues, we propose a
novel non-autoregressive framework, named \emph{Fast Neural Image Caption} (FNIC), which integrates the position information to guide the generation of descriptions.
Specifically, as shown in Figure \ref{fig1},  we first adopt an image encoder, which extracts feature maps from CNNs or object features from object detection models, then maps the dimension of input visual features to the model dimension of the decoder using a multi-layer perceptron (MLP). As for the decoder, we utilize position alignment to generate and order the words that roughly describe the content of the image, and then fine the organized words into the final sentence. We further introduce a decoding inference strategy which adopts position information to guide the sentence generation.
Rather than just base on roughly generated words, we utilize the conditional distribution of these ordered words as a latent variable to guide the decoding of the target sentence. Intuitively, with the guide of position information, for each target word, what it is expressed can be nearly limited to the corresponding word of coarse ordered words in the same position.

The experimental results on different widely-used public benchmarks show that our proposed captioning model achieves significant and consistent improvements compared to general non-autoregressive models through explicitly adopting the position information to guide the decoding. Moreover, by introducing a simple but effective auto-regressive decoder to construct positive information and generate coarse words, our model immensely narrows the description quality gap between autoregressive and NA sentence generation models, while maintains a considerable speedup (nearly 8 times faster). We will release all source codes and related resources of this work for further research explorations.

Our contributions are summarized as follows:
\begin{itemize}
	\item 
	We introduced non-autoregressive technology into the captioning task, which effectively accelerates the description generation while maintains a comparable performance.
	\item To improve the quality of generated sentences and overcome the semantic gap between image and text, we design a light position alignment module based on object detection features to produce ordered words.
	\item Rather than just base on roughly generated ordered words, we utilize the conditional distribution of these ordered words as a latent variable to guide the decoding of the target sentence.
\end{itemize}

\section{Background}
\subsection{Image Caption}
In an image captioning task, a stand encoder-decoder framework first encodes the given image ${I}$ to some visual feature vectors ${I}_F$ and then generates descriptive sentence ${S}$ according to previously generated words and image features \cite{Vinyals2015Show,Fang2015From,Donahue2015Long,Lu2016Visual}. It corresponds to the word-by-word nature of human language generation and effectively learns the conditional distribution of real captions. Formally, autoregressive models consider the distribution over possible output sentences ${S}$ as a chain of conditional probabilities,
\begin{equation}
P({S}|{I}_F)=\prod_{i=1}^{n}P(w_i|{w}_{<i},{I}_F)
\end{equation}

In the training procedure, the model parameters are trained to minimize the conditional cross-entropy loss, where the loss function is defined as: 
\begin{equation}
\mathcal{L}_A=-\sum_{i=1}^{n} \text{log} \ P(w_i|{w}_{<i}, {I}_F)
\end{equation}

During inference procedure, at each step, a word is produced and then fed into the decoder to predict the next word. Therefore, all autoregressive decoders must remain sequential rather than parallel when generating sentences. Moreover, sequential decoding
is tend to copy words from the training data to enhance the grammatical accuracy, which easily causes semantic error and limits the diversity of descriptions \cite{Jiang2018Recurrent}.

\subsection{Non-autoregressive Decoder}
Non-autoregressive neural decoder is first proposed by \cite{GuNon} to solve the slow generation issue in autoregressive neural machine translation, which could simultaneously generate target words by removing their dependencies. Formally,
\begin{equation}
P({S}|{I}_F)=\prod_{i=1}^{n}P(w_i|{I}_F)
\end{equation}
Instead of utilizing the previous subsentence history, NA models usually directly copy the source feature vectors as the input of the decoder. Hence, when generating a sentence, NA decoder could predict all target words with their maximum likelihood individually by breaking the dependency among the target words, and therefore
the decoding procedure of NA models is in parallel and gains very fast generation speed.

However, since NA decoder discards the sequential dependencies among words in the target sentence, it suffers from the potential performance degradation due to the limited information for decision making. To be specific, when decoding
a target word, the NA model must be able to figure out not only what target-side information does the word describe but also what is expressed by other target words \cite{Bingzhen2019Imitation,Guo2018Non}. 
With the limited information to make predictions, NA models cannot effectively learn the intricate patterns from source information to target sentences, which leads to inferior language quality. Therefore, the position alignment module is added to increase the location information and reduce the candidate search space.

\section{Method}

In this section, we detail our proposed FNIC architecture which integrates position alignment structure into decoder, pursuing an improved non-autoregressive sentence generation to accelerate image captioning. We first describe the popular image encoder types,  then show how we implement a hierarchical decoder which consists of a position alignment and a fine sentence decoder, to realize a non-autoregressive decoding procedure. Finally, we introduce a non-deterministic probability inference process as well as the model training process.

\subsection{Image Encoder}

The image encoder in general encoder-decoder framework aims to extract a set of imgae feature vectors ${I}_F=\{v_1,v_2,\ldots,v_{k}\}$ for different regions based on given image {I}, where $k$ is the number of regions. A typical image encoder usually adopts a CNN (\emph{e.g.} ResNet \cite{He2016Deep}) to extract features. Moreover, R-CNN based models (\emph{e.g.} Faster RCNN \cite{ShaoqingFaster}) are employed to improve the captioning performance which utilizes bottom-up attention \cite{Anderson2017Bottom} and  provides a better understanding of objects in the image.

To be specific, for a CNN based encoder, a spatially adaptive max-pooling layer is adopted after a classical CNN structure to fix the size of the output vector and streamlining information.
As for an R-CNN based encoder, which usually detects objects in two stages. The first stage predicts region proposals, and the second stage predicts class labels as well as bounding box refinements. 
To obtain a feature vector set, we select all regions in which any class detection probability exceeds a confidence threshold and apply mean-pooling on the convolutional
features for each of vectors. 

In most cases, we utilize a linear transformation to project all image feature vectors in ${I}_F$  to a fixed size of $d$ before input to the decoder module.

\subsection{Hierarchical Decoder}
Here, we will introduce a novel non-autoregressive decoder, which aims to accelerate sentence generation while keeps comparable performance.
As shown in Figure \ref{fig1}, our proposed FNIC employs a position alignment module to explicitly model the position information in the decoding. Formally, FNIC first transforms the image feature ${I}_F$ into a coarse ordered words ${X}=\{x_1,x_2, \ldots, x_m\}$ which the order of coarse words appears is the same as that in the target sentence, and then expands the ordered words to a target sentence ${S}$. Our proposed decoder handle  the overall image captioning probability as:
\begin{equation}
P({S}|{I}_F)=\sum_{{X}}P({X}|{I}_F)P({S}|{X},{I}_F)
\end{equation}
where $P({X}|{I}_F)$ is realized by the position alignment module and $P( {S}|{X},{I}_F)$ is implemented by the fine decoder module. 

\subsubsection{Position Aligment}

The position alignment module determines the order of words corresponding to the target sentence, which represents the objects and scenes in given image ${I}$,  by learning to transform the image feature ${I}_F$ into the ordered words ${X}$.

Since we conduct an experiment and find that autoregressive models are more suitable for modeling the position information compared to NA models, and even a light autoregressive model with similar decoding speed to a large NA model could achieve better performance in modeling position information. Hence, we  integrate a light autoregressive model to
build the coarse ordered words probability as:

\begin{equation}
P({X}|{I}_F)=\prod_{i=1}^{m}P(x_i|{x}_{<i},{I}_F)
\end{equation}
where ${x}_{<i}=\{x_1,x_x,\ldots,x_{i-1}\}$ represents the ordered word history.

In practical terms, we use a greedy search algorithm to generate ordered words word-by-word. When decoding the $i$-th ordered word $x_i$, we obtain its hidden representation as:
\begin{equation}
\text{R}_i=\text{GRU}(\text{R}_{i-1}|[{C}_{i-1};{X}_{<i}])
\end{equation}
where \text{GRU}($\cdot$) is a one-layer gated recurrent unit \cite{Cho2014Learning} and ${C}_{i-1}$ is calculated by the inter-attention to the extracted image features. After obtaining the hidden representation, the conditional probability can be calculated as,
\begin{equation}
P(x_i|{I}_F)=\text{softmax}(\text{R}_i)
\end{equation}

\subsubsection{Fine Decoder}
The fine decoder module generates the target sentence with the guiding of coarse ordered words, which regards the generation of each word as non-autoregressive:

\begin{equation}
P({S}|{X},{I}_F)=\prod_{i=1}^{n}P(w_i|{X},{I}_F)
\end{equation}

Different from the original NA decoder, the inputs of our decoder module is the ordered words plus image features instead of just copied source information, which is used to guide the decoding direction.

In actual work, we employ an NA model with $N$ layers of Transformer decoder which is composed of a multi-head self-attention layer, a multi-head inter attention layer and a feed-forward layer.
\begin{equation}
D({Y},{I}_F)=\text{FFN}(\text{ATT}_{inter}({I}_F,\text{ATT}_{self}({Y})))
\label{eq:9}
\end{equation}
where FFN($\cdot$) is the feed-ward layer and ${Y}$ is the input of fine decoder.

\subsection{Inference Procedure}
Our FNIC model explicitly builds position information of non-autoregressive and aims to utilize it to provide effective limitation information and assist the generation of target sentences. Now the remaining problem is how to perform decoding with the guide of ordered word information.
We propose to utilize the coarse ordered words as a bridge to guide the decoding of the target sentence, which can be formulated as:
\begin{equation}
\begin{split}
{S}^*&=\text{argmax}_{{S}}P({S}|{I}_F)\\
&=\text{argmax}_{{S}}\sum_{{X}}P({S}|{X},{I}_F)P({X}|{I}_F)
\end{split}
\label{eq:6}
\end{equation}

It is impractical to obtain a direct and accurate solution for maximizing Equation \ref{eq:6} due to the high time complexity. 
One easy way to think about it is that first generates the most probable ordered words based on image features and then decodes the target sentence conditioned on it:
\begin{gather}
{X}^*=\text{argmax}_{{X}}P({X}|{I}_F) \\
{S}^*=\text{argmax}_{{S}}P({S}|{X}^*,{I}_F) 
\end{gather}
This two-stage local optimal approach is simple and effective, but it brings in some noise in the approximation and can not guarantee a global optimum solution.

Inspired from \emph{Hidden Markov Model} (HMM),  
different from the local deterministic optimal strategy which utilizes a deterministic ordered word set  to guide,  we introduce a new inference strategy, regards the probability distribution ${Q}$ of the ordered words as a latent variable, and models the decoder procedure as generating the target sentence according to the latent variable ${Q}$, \emph{i.e.}, Equation \ref{eq:6} is re-formulated as:
\begin{equation}
{S}^*=\text{argmax}_{{S}}P({S}|{Q},{I}_F)
\end{equation}
where the probability distribution ${Q}$ is defined as:

\begin{equation}
{Q}({X})=P({X}|{I}_F)=\dfrac{\text{exp}(f({X}))}{\sum_{{X}'}\text{exp}(f({X}'))}
\end{equation}
where $f(\cdot)$ is a score function of ordered words (the input of the softmax layer in the fine decoder). Since the latent variable ${Q}$ can be viewed as a non-deterministic form of the guiding ordered words, the
sentence generation with the non-deterministic strategy is also guided by the hidden ordered words.

To be specific,  we use the weighted word embeddings of the conditional probability ${Q}$ as the input of the decoder, then obtain the final probability:
\begin{equation}
P({S}|{Q},{I}_F)=\text{softmax}(D^{N}({Q}^T\text{Emb}({S}),{I}_F))
\end{equation}
where $D(\cdot)^N$ is $N$-layer fine decoder in Equation \ref{eq:9}.
Please note that the major difference between local deterministic and non-deterministic   probability strategy is the inputs of fine decoder module (Location \ding{173} in Figure \ref{fig1}), where the local deterministic strategy directly utilizes the word embeddings of generated coarse ordered words and the non-deterministic   probability  strategy utilizes the weighted word embeddings of the word probability of position information.

\subsection{Training Procedure}

In the training process, for each training sentence pair $({I, S})$ in data set $\mathbb{D}$, we first utilize image encoder to extract salient object labels, and then 
generate its corresponding candidate ordered word set $\hat{{X}}$: we adopt a word position alignment tool to select the words that appear in both the target sentence ${I}$ and the object labels, and then sort the selected words according to the order in the target sentence.
At last, FNIC is optimized by maximizing a joint loss:

\begin{equation}
\mathcal{L}=\mathcal{L}_P+\mathcal{L}_F
\end{equation}
where $\mathcal{L}_P$ and $\mathcal{L}_F$ represent the position alignment and sentence generation losses respectively. Formally, based on non-deterministic probability approach, the position alignment loss $\mathcal{L}_P$ is defined as:

\begin{equation}
\mathcal{L}_P=\sum_{({I},\hat{{X}},{S})\in \mathbb{D}} \text{log}P(\hat{{X}}|{I}_F)
\end{equation}
And the fine sentence generation loss $\mathcal{L}_F$ is defined as an overall maximum likelihood of decoding the target sentence from the conditional probability of ordered words:
\begin{equation}
\mathcal{L}_F=\sum_{({I},\hat{{X}},{S})\in \mathbb{D}} \text{log}P({S}|{Q},{I}_F)
\end{equation}

In particular, we use the trained model for the local deterministic optimal approach to initialize the model before the non-deterministic latent variable approach, since if ${Q}$ is not well trained, $\mathcal{L}$ will converge very slowly.

\section{Experiments}

\subsection{Dataset}

\begin{table*}[!h]
	\begin{center}
			{\caption{Performance comparisons with different evaluation metrics on the COCO Karpathy test split. All values except Speed up are reported as percentage (\%).}
			\label{tab1}}
		\begin{tabular}{lcccccc}
			\toprule
			&BLEU@4&METEOR&ROUGE-L&CIDEr-D&SPICE&Speed up\\
			\midrule
			SCST \cite{Rennie2017Self} &30.0&25.9&53.4&99.4&-&- \\
			ADP-ATT \cite{Lu2016Knowing} &33.2&26.6&-&108.5&-&- \\
			LSTM-A \cite{Yao2016Boosting} &35.2&26.9&55.8&108.8&20.0&-\\
			Up-Down \cite{Anderson2017Bottom}  &36.2&27.0&56.4&113.5&20.3&-\\ 
			GCN-LSTM \cite{Yao2018Exploring} &\textbf{37.0}&\textbf{28.1}&\textbf{57.1}&\textbf{117.1}&\textbf{21.1}&-\\ \hline
			NAIC  &28.5&23.6&52.3&98.2&18.5&\textbf{12.40}$\times$\\ 
			FNIC$_{\text{AT}}$  &36.6&27.2 & 56.2&118.0&20.5&1.00$\times$\\ 
			FNIC$_{\text{NAT}}$  &30.4&25.4&55.1&107.4&19.6&11.21$\times$\\ 
			FNIC&36.2&27.1 & 55.3&115.7&20.2&8.15$\times$\\
			\bottomrule
		\end{tabular}
	\end{center}
\end{table*}

\textbf{COCO} is a standard benchmark for image captioning task which contains 123,287 images (82,783 for training and 40,504 for validation) and each image is annotated with 5 descriptions by humans. Since the annotated descriptions of the official testing set are not provided, we utilize Karpathy split (113,287 for training, 5,000 for validation and 5,000 for testing) as in \cite{Anderson2017Bottom}. According to \cite{Karpathy2016Deep}, all the training sentences are converted to lower case and we omit rare words which occur less than 5 times. Therefore, the final vocabulary includes 10,201 unique words.\\

\noindent
\textbf{Visual Genome}, which contains images with annotated objects, attributes, and relationships, is adopted to train Faster R-CNN for object detection. In this paper, we follow the setting in \cite{Anderson2017Bottom,Yao2018Exploring} and take 98,077 images for training, 5,000 for validation, and 5,000 for testing. As in \cite{Anderson2017Bottom}, 1,600 objects and 400 attributes are considered from Visual Genome for training Faster R-CNN with two branches for predicting objects and attribute classes.

\subsection{Baselines}

Since the non-autoregressive method in the image captioning task has been underinvestigated so far, there are few existing baselines for our comparison. We choose to compare with the following typical autoregressive models for their high relevance with our task as well as their superior performance demonstrated by previous works:

(1) \textbf{SCST} \cite{Rennie2017Self} employs a self-critical sequence training strategy to train a modified visual attention-based captioning model in \cite{Xu2015Show}  (2) \textbf{ADP-ATT} \cite{Lu2016Knowing} develops an adaptive attention based encoder-decoder model for automatically determining whether to attend
to the image and which image regions to focus for caption. (3) \textbf{LSTM-A} \cite{Yao2016Boosting} integrates semantic attributes into CNN plus RNN captioning model for boosting image captioning. (4) \textbf{Up-Down} \cite{Anderson2017Bottom} designs a combined bottom-up and top-down attention mechanism that enables region-level attention to be calculated to boost image caption. (5) \textbf{GCN-LSTM} \cite{Yao2018Exploring} further exploits visual relationships between objects through graph convolutional networks. 

We also select three models as baselines: (1) \textbf{NAIC} develops only non-autoregressive fine decoder (one-layer Transformer decoder) without position alignment. (2)  $\textbf{FNIC}_{\text{AT}}$ adopts autoregressive position alignment (one-layer GRU) and autoregressive fine decoder (one-layer  Transformer decoder). (3)  $\textbf{FNIC}_{\text{NAT}}$ utilizes both non-autoregressive position alignment (one-layer  Transformer decoder) and non-autoregressive fine decoder (one-layer Transformer decoder). The model architectures of $\textbf{FNIC}_{\text{AT}}$, $\textbf{FNIC}_{\text{NAT}}$ and our proposed $\textbf{FNIC}$ are identical. 
Note that for fair comparison, all the baselines and our model adopt ResNet-101 as the basic architecture of image feature extractor. All models use the same training strategy and are evaluated using greedy decoding. 

\subsection{Implementation Details}

For each image, we apply Faster R-CNN to detect objects within this image and select top $K$ = 36 regions with highest detection confidences to represent the image. The dimension of each region is set as 2,048. 
For fine decoder module, we utilize a one-layer Transformer model ($D_{model}$=512, $D_{hidden}$=512, $n_{head}$=2, $dropout$=0.1). For the GRU position alignment module, we set it to have the same hidden size with the Transformer model.
The captioning model is mainly implemented with PyTorch (version 1.1.0), optimized with Adam \cite{Kingma2014Adam}. We set the initial learning rate as 0.0005 and the mini-batch size as 1,024. The momentum and the weight-decay are 0.8 and 0.999 respectively.  The maximum training iteration is set as 35 epochs. 
In addition, we write a position ranking tool based on python (version 3.6) to rank the object labels appear in the target sentence.

\subsection{Automatic Evaluation}

\subsubsection{Metrics}
Following previous image captioning work \cite{Yao2018Exploring,Anderson2017Bottom}, 
we select five types of metrics: BLEU@N \cite{Papineni2002BLEU}, METEOR \cite{Lavie2007METEOR}, ROUGE-L \cite{Flick2004ROUGE}, CIDEr-D \cite{Vedantam2015CIDEr} and SPICE \cite{Anderson2016SPICE}. In particular, SPICE focuses on semantic analysis and has a higher correlation with human judgment, and other metrics favor frequent training n-grams and measure the overall sentence fluency, which are more preferred in sequential decoding based methods.  For all the metrics, the larger they are, the more relevant or fluent the generated descriptions are.
 Speed up is computed
based on the time to decode a single sentence without minibatching, and the values are averaged over the whole offline test set. The decoding is implemented on a single NVIDIA GeForce GTX 1080 Ti.
\begin{figure*}[t]
	\begin{center}
		{\rule{0pt}{2in}
			\includegraphics[width=0.9\linewidth]{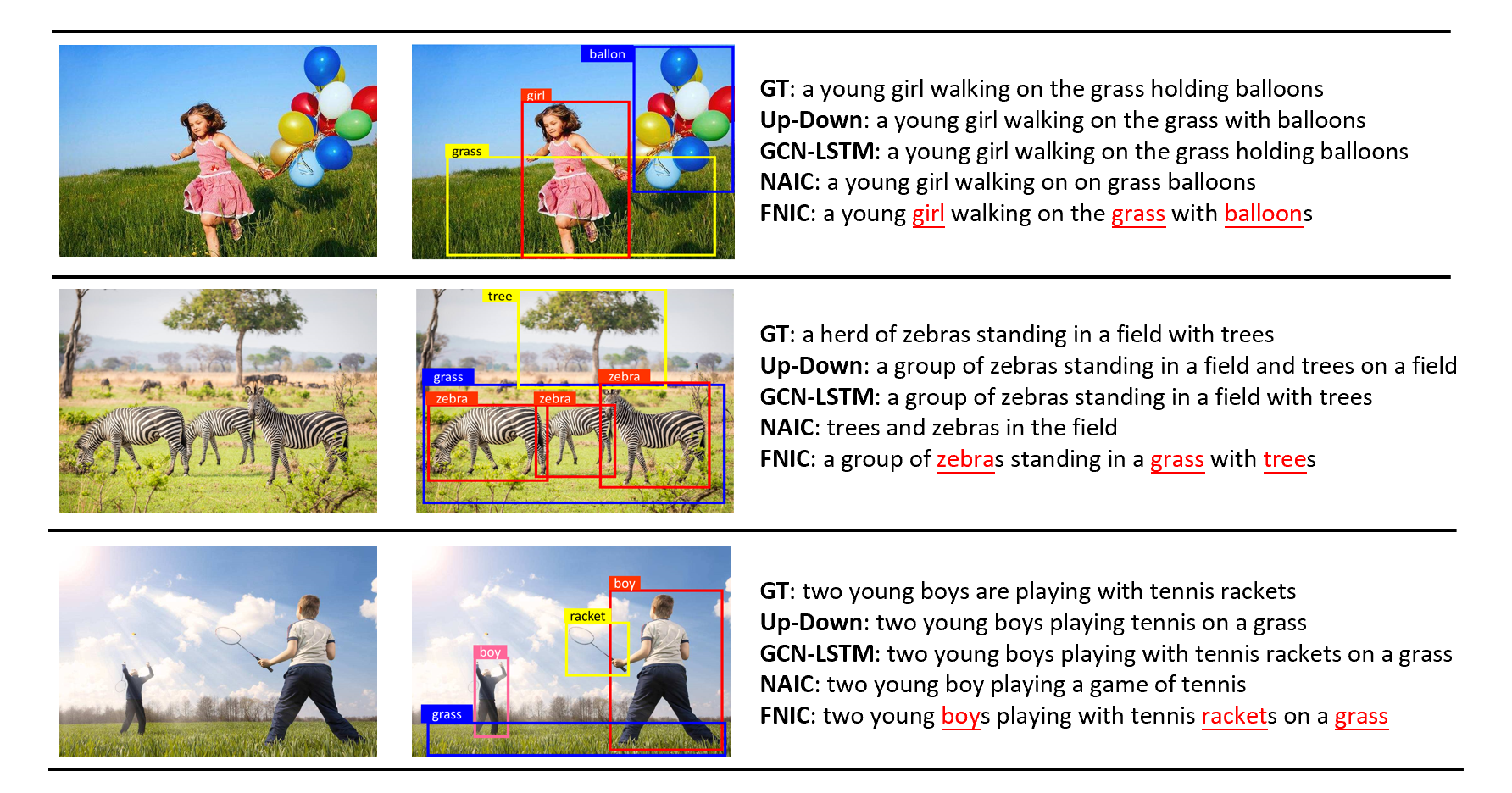}
		}
	\end{center}
	\caption{Image examples from \cite{Lin2014Microsoft} with object regions and sentence generation results. The output
		sentences are generated by (1) Ground Truth (GT): One ground truth sentence, (2) Up-Down (3) GCN-LSTM (4) NAIC and (5) our proposed FNIC.}
	\label{fig4}
\end{figure*}

\begin{table*}
	\begin{center}
			{\caption{Effect of non-deterministic inference strategy on COCO Karpathy test split. DT represents local deterministic inference and NDT represents non-deterministic probability inference. All values are reported as percentage (\%).}
		\begin{tabular}{lccccc}
			\toprule
			&BLEU@4&METEOR&ROUGE-L&CIDEr-D&SPICE\\
			\midrule
			FNIC$_{\text{NAT}}$+DT &29.0&24.9&53.4&106.2&19.2\\
			FNIC$_{\text{NAT}}$+NDT &30.4&25.4&55.1&107.4&19.6\\
			FNIC+DT &35.6&26.6&53.2&113.5&19.5 \\
			FNIC+NDT &\textbf{36.2}&\textbf{27.1}&\textbf{55.3}&\textbf{115.7}&\textbf{20.2}\\
			\bottomrule
		\end{tabular}
		\label{tab2}}
	\end{center}
\end{table*}

\subsubsection{Overall Results and Analysis}
We compare our proposed model {Fast Neural Image Caption} (FNIC) with all
baseline models. All the results are shown in Table \ref{tab1}. From the table, we can find that:

(1) FNIC achieves excellent performance on most of the benchmark metrics,
which is even close to the best AT model with smaller
than 1 BLEU gap ($36.2$ vs. $37.0$). It is also worth mentioning that although
FNIC utilizes a small GRU model to better capture position information, it could still maintain low sentence generation latency (about 8 speedup of \textbf{FNIC}$_{\text{AT}}$ which only utilizes autoregressive decoder). 
Compared to \textbf{NAIC}  which does not keep the position alignment module, our \textbf{FNIC} provides effective position and semantic information in the generated coarse ordered words, which limit the candidate word space in the source sentence generation and makes it faster. Another reason is that a major potential performance degradation of non-autoregressive models compared to autoregressive models comes from the difficulty of modeling the sentence structure difference between source information 
and target sentences, \emph{i.e.}, position and semantic information, which is neglected for most of existing non-autoregressive models but can be well modeled by the small GRU position alignment decoder.

(2) A light GRU position alignment model with close latency to large
non-autoregressive models could perform much better in modeling position information. On all benchmark metrics, \textbf{FNIC} with a small autoregressive GRU position alignment module achieves a much better description quality than that with a large non-autoregressive model (one-layer Transformer decoder has a larger parameter scale than one-layer GRU). Moreover, we find that our proposed model \textbf{FNIC} with a
one-layer GRU for decoding could even outperform some of the existing autoregressive models such as \textbf{SCST}, \textbf{ADP-ATT} and \textbf{LSTM-A} in most metrics, while maintains acceptable latency. These verify that ordered words modeled bt position alignment module could effectively reduce the decoding space and improve the generated sentence quality of the model.

\subsection{Effect of Non-deterministic Inference Strategy}

We also investigate the effect of two proposed
inference strategies including local deterministic and non-deterministic probability on COCO Karpathy test split. In Table \ref{tab2}, we can find that the non-deterministic probability strategy has better performance compared to the local deterministic inference strategy for both \textbf{FNIC} and \textbf{FNIC}$_{\text{NAT}}$ since the non-deterministic probability strategy could effectively reduce the information loss of the deterministic strategy.
On the other hand, we can also find that the non-deterministic probability strategy does not bring many improvements for our best model with the GRU position alignment module. One reason is perhaps that the coarse ordered words generated by the GRU position alignment module are good enough and therefore it does not bring in much information loss in the whole decoding procedure. Therefore, we use non-deterministic inference as the default decoding strategy in the proposed model.

\subsection{Case Study}

Figure \ref{fig4} shows example generated descriptions  of top-ranking autoregressive model \textbf{Up-Down}, \textbf{GCN-LSTM}, general non-autoregressive  captioning model \textbf{NAIC} and our proposed model \textbf{FNIC}. We find that
the problem of missing objects and repeated descriptions are severe in the general NA model (Such as repeated preposition “on” in the second image and missing object “grass” in third image), while this problem is effectively alleviated in \textbf{FNIC}
model. Moreover, we find that most of the missing, or wrong word in the generated sentence of general NA model comes from the errors in the coarse ordered words, which demonstrates that the NA model could well transform the coarse ordered words which
have the target sentence structure to the final fine decoder.

\subsection{Diversity Study}

\begin{table}
	\begin{center}
		{\caption{Illustration of the diversity of different methods from different aspects. All values are reported as percentage (\%).}
			\label{tab3}}
		\begin{tabular}{lccc}
			\toprule
			&Novel&Unique&Vocabulary Usage\\
			\midrule
			Up-Down &45.05&61.58&7.97\\
			GCN+LSTM &60.23&83.22&8.64\\
			FNIC$_{\text{AT}}$ &55.12&78.42&8.33 \\
			FNIC &\textbf{81.25}&\textbf{87.12}&\textbf{12.16}\\
			\bottomrule
		\end{tabular}
	\end{center}
\end{table}
Diversity is an important indicator of image captioning tasks.
Since the non-autoregressive decoding approach is a visual-to-linguistic transformation process and can provide more semantic information, it is distinctive to autoregressive decoding that more diverse captions are expected to be generated. 

To analyze the diversity of generated captions, according to \cite{DaiA},
we compute \textbf{novel} caption percentage, \textbf{unique} caption percentage and \textbf{vocabulary usage}, which respectively account for the percentage of captions that have not been seen in the training data, the percentage of captions that are unique in the whole generated captions and the percentage of words in the vocabulary that are adopted to generate captions. According to Table \ref{tab3}, it is obvious that \textbf{FNIC} can achieve better results, which suggest that NA methods can generate more
diverse captions than sequential methods, such as \textbf{TopDown } and \textbf{FNIC}$_{\text{AT}}$.

\section{Conclusion}
In this paper,  we find that the limited information for each word generation is a important factor leading to the poor performance of NA decoding. To cover this problem, we propose a fast  and effective image captioning framework named FNIC which explicitly integrates the extra information through a light GRU position alignment module in the decoding procedure. 
Furthermore, we introduce non-deterministic probability inference strategies to utilize the coarse ordered words to narrow the candidate decoding search space and guide the fine sentence generation in the next step. 
Experimental results on public datasets
show that our NA model achieves better performance than general NA captioning models, and  keeps  comparable sentence quality as popular
autoregressive model while keeps a significant speedup. We believe
that well model the position and semantic information is a potential way for better NA in the future.

\section{Acknowledgments}
We would like to thank Haomiao Sun for helpful discussions.

\bibliographystyle{aaai} 
\bibliography{rerank} 
\end{document}